\documentclass{Interspeech}

\interspeechcameraready

\title{Calm-Whisper: Reduce Whisper Hallucination On Non-Speech By Calming Crazy Heads Down}

\author[affiliation={1}]{Yingzhi}{Wang}
\author[affiliation={1}]{Anas}{Alhmoud}
\author[affiliation={1}]{Saad}{Alsahly}
\author[affiliation={1}]{Muhammad}{Alqurishi}
\author[affiliation={2,3}]{Mirco}{Ravanelli}

\affiliation{Research Center}{Elm Company}{KSA}
\affiliation{}{Concordia University}{Canada}
\affiliation{}{Mila}{Canada}
\email{ywang@elm.sa, aalhmoud@elm.sa, salsahly@elm.sa, mualqurishi@elm.sa, mirco.ravanelli@concordia.ca}


\usepackage{comment}

\begin{document}

\maketitle

\begin{abstract}
OpenAI's Whisper has achieved significant success in Automatic Speech Recognition. However, it has consistently been found to exhibit hallucination issues, particularly in non-speech segments, which limits its broader application in complex industrial settings. 

In this paper, we introduce a novel method to reduce Whisper's hallucination on non-speech segments without using any pre- or post-possessing techniques. Specifically, we benchmark the contribution of each self-attentional head in the Whisper-large-v3 decoder to the hallucination problem by performing a head-wise mask. Our findings reveal that only 3 of the 20 heads account for over 75\% of the hallucinations on the UrbanSound dataset. We then fine-tune these three crazy heads using a collection of non-speech data. The results show that our best fine-tuned model, namely Calm-Whisper, achieves over 80\% reduction in non-speech hallucination with only less than 0.1\% WER degradation on LibriSpeech test-clean and test-other.
\end{abstract}

\section{Introduction}

OpenAI's Whisper \cite{whisper} has emerged as a significant milestone in Automatic Speech Recognition (ASR), owing to its multilingual capabilities, high transcription accuracy, and adaptability to diverse acoustic environments. The model is built on a large-scale Transformer-based encoder-decoder architecture, boasting billions of parameters that allow it to capture complex details in speech signals. Whisper's training leverages an extensive dataset comprising 680k hours of multilingual and multitask speech data. The diversity of the dataset ensures that the model can generalise effectively to different scenarios. 

However, like many large-scale generative models, Whisper has been shown to exhibit hallucination issues, where it generates text that is unrelated to the original audio input  \cite{h1, h2, h5, h7, hallucination_1}. These hallucinations, manifesting as fabricated phrases or entire sentences, raise concerns about the reliability and fairness of ASR systems in real-world applications \cite{careless}.
Key factors contributing to Whisper's hallucinations can be categorized into three perspectives: model design, training data quality, and input ambiguity \cite{hallucination_0}. Whisper's large-scale transformer-based architecture, while highly generative, can unintentionally generate plausible yet incorrect outputs, with its auto-regressive nature further amplifying its hallucination tendencies. The training corpus of Whisper includes weakly annotated data that may be biased, imbalanced, or noisy, leading to an increasing risk of hallucinations \cite{careless}. As a result, when the input is noisy, unclear, or contains ambiguous contexts, Whisper is highly prone to produce hallucinations.

Previous studies have explored Whisper's hallucination problem and offered potential solutions. 
In Careless-Whisper \cite{careless}, the authors highlight that approximately 1\% of Whisper's transcriptions contain hallucinated phrases or sentences not present in the original audio. Notably, 38\% of these hallucinations include explicit harms such as perpetuating violence, making inaccurate associations, or implying false authority. They also find that hallucinations disproportionately occur for individuals with speech impairments, like aphasia, especially when there are longer non-vocal durations in speech.
In Distil-Whisper \cite{distil-whisper}, the authors used a simple word error rate (WER) heuristic to select only the highest quality pseudo-labels for training. The study notes that Distil-Whisper is less prone to hallucination errors on long-form audio compared to the original Whisper model.
WhisperX \cite{whisperx} proposes pre-processing operations involving Voice Activity Detection (VAD), and proves that these VAD operations reduce hallucination and repeating on the Kincaid46 \cite{Kincaid4} and TED-LIUM \cite{ted} benchmarks.

Unlike the speech environment in the above-mentioned studies, industrial environments are filled with complex acoustic conditions, including audio segments containing only non-speech noise or varying lengths of silence. The Whisper model's lack of robustness in handling such inputs has been long criticized\footnote{\url{https://github.com/openai/whisper/discussions/1606}}. 
While hallucination on speech content can lead to misinformation, in most cases, such hallucinations are recognizable and can be filtered out by reviewers. Despite misinformation, the primary content of the speech remains understandable, allowing users to grasp the main ideas. In contrast, hallucinations on non-speech content pose a greater challenge by fabricating non-existent inputs. This can result in significant confusion in understanding transcriptions, as users may struggle to distinguish whether the entire segments of transcription are real.
In \cite{non-speech}, the authors also identified the hallucination issue on non-speech inputs and attempted to remove it by applying VAD as pre-processing and an n-gram language model as post-processing. However, these techniques remain limited in effectiveness.
Although VAD can be used to remove non-speech segments, due to the inherent limitations of VAD models, complete removal is unachievable. For example Silero VAD v5 \cite{silero} achieves only 61\% utterance-level accuracy on the ESC-50 dataset\cite{esc-50}, suggesting that up to 40\% of pure noise in similar scenarios could be misinterpreted as speech and propagate through the recognition pipeline. Regarding post-processing, using an additional language model as filter \cite{non-speech} may also filter out real speech contents, causing a loss of information.

In this paper, we seek to eliminate Whisper's non-speech hallucinations within the model itself.
We hypothesize that only certain self-attention heads in the decoder are noise-sensitive and hallucinatory. To validate this, we mask these heads one by one and benchmark their impact on hallucination occurrence. After identifying the most hallucinatory heads, we conduct calm-down fine-tuning only on these heads using non-speech audio paired with blank labels. Results show that our approach reduces non-speech hallucinations by over 80\% while maintaining undegraded speech recognition accuracy.

\begin{figure}[t!]
  \centering
  \includegraphics[width=\linewidth]{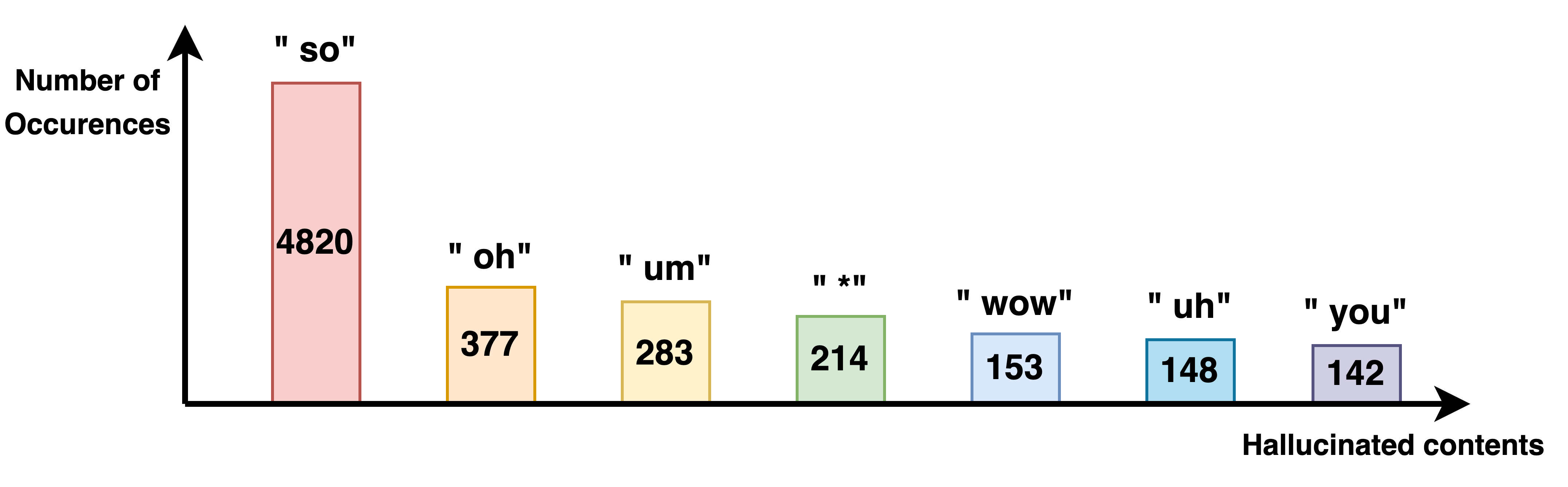}
  \caption{Most frequently hallucinated sentences by Whisper-large-v3 on UrbanSound8K}
  \label{fig:content}
\end{figure}

\section{Whisper Hallucination On Non-Speech}
As a first step, we run Whisper-large-v3\footnote{\url{https://huggingface.co/openai/whisper-large-v3}} on the UrbanSound8K dataset \cite{urbansound} to showcase its significant hallucination problem dealing with non-speech inputs.

\subsection{Evaluation Set}
UrbanSound8K is a widely used environmental sound dataset designed for audio classification and recognition tasks. It consists of 8,732 labeled sound clips spanning 10 different classes of urban sounds, including street music, air conditioners, and other noise sources that could be commonly encountered in industrial settings. Each audio clip is up to four seconds long, which is also similar to the typical pause duration of speakers in industrial environments such as call centers.

\subsection{Metric}
For speech content, the insertion error rate (IER) is commonly used as a metric to measure a model's hallucination. However, for non-speech content, where the total number of words in the sentence is zero, IER cannot be calculated. In such case, we abandon word-level evaluation methods and adopt a stricter utterance-level metric. We consider the hallucination assessment for non-speech content as a binary classification task: for an entire audio clip, the model's output is categorized as either hallucinating or not hallucinating. A robust model is expected to produce an empty transcription for non-speech input, which is considered "not hallucinating," whereas any non-zero-length transcription is regarded as "hallucinating". Therefore, the hallucination rate, which quantifies non-speech hallucination, can be defined as:
\begin{equation}
  Hallucination\ Rate = \frac{N_{len(transcription)>0}}{N_{total}}
\end{equation}
where $N_{len(transcription)>0}$ refers to the number of hallucinating transcriptions and $N_{total}$ is the total number of audios in the dataset.
Throughout the rest of this paper, we employ it as the standard metric for assessing non-speech hallucinations.

\subsection{Results}
For comparison, we also run a popular CTC \cite{ctc} model Conformer-CTC-Large\footnote{\url{https://catalog.ngc.nvidia.com/orgs/nvidia/teams/nemo/models/stt_en_conformer_ctc_large}} \cite{conformer}. As is shown in Table 3, Whisper-large-v3 exhibits an enormous hallucination rate of nearly 100\%, whereas Conformer-CTC is considerably more robust, with a hallucination rate of only 13.52\%.

\begin{table}[h!]
  \caption{Hallucination Rate of Whisper-large-v3 and Conformer-CTC-large on UrbanSound8K.}
  \label{tab:hal_rate}
  \centering
  \resizebox{\columnwidth}{!}{
  \begin{tabular}{c c c}
    \toprule
     & Whisper-large-v3 & Conformer-CTC-large \\
    \midrule
    Hallucination Rate & 99.97\% & 13.52\% \\
    \bottomrule
  \end{tabular}
  }
\end{table}


Additionally, in Figure 1, we present some of the most frequently hallucinated contents. It can be observed that Whisper-large-v3 tends to misinterpret non-speech sounds as filler words or markers of tone and pause, with as much as 55.2\% of the audios being transcribed into \textit{" so"}. Moreover, we found that 8.5\% of the hallucinations span more than five tokens, which might significantly disrupt the overall comprehension of the transcription under industrial conditions.

\section{Hallucinatory Heads Detection}
In \cite{retrieval_head}, the authors investigated the role of the "retrieval head" in transformer-based language models in maintaining factual accuracy over long contexts. Inspired by this research, we hypothesize that the decoder in the Whisper model, which handles language modeling, may also have "crazy heads" that are more prone to causing hallucinations. To verify this effect, we conduct experiments using both single-head masking and multi-head masking approaches.

\subsection{Single-Head Masking}
We begin by masking each of the 20 self-attentional heads (\#0-\#19) in the decoder, one at a time, and subsequently measure the hallucination rate on the UrbanSound8k dataset.

\begin{figure}[t!]
  \centering
  \includegraphics[width=\linewidth]{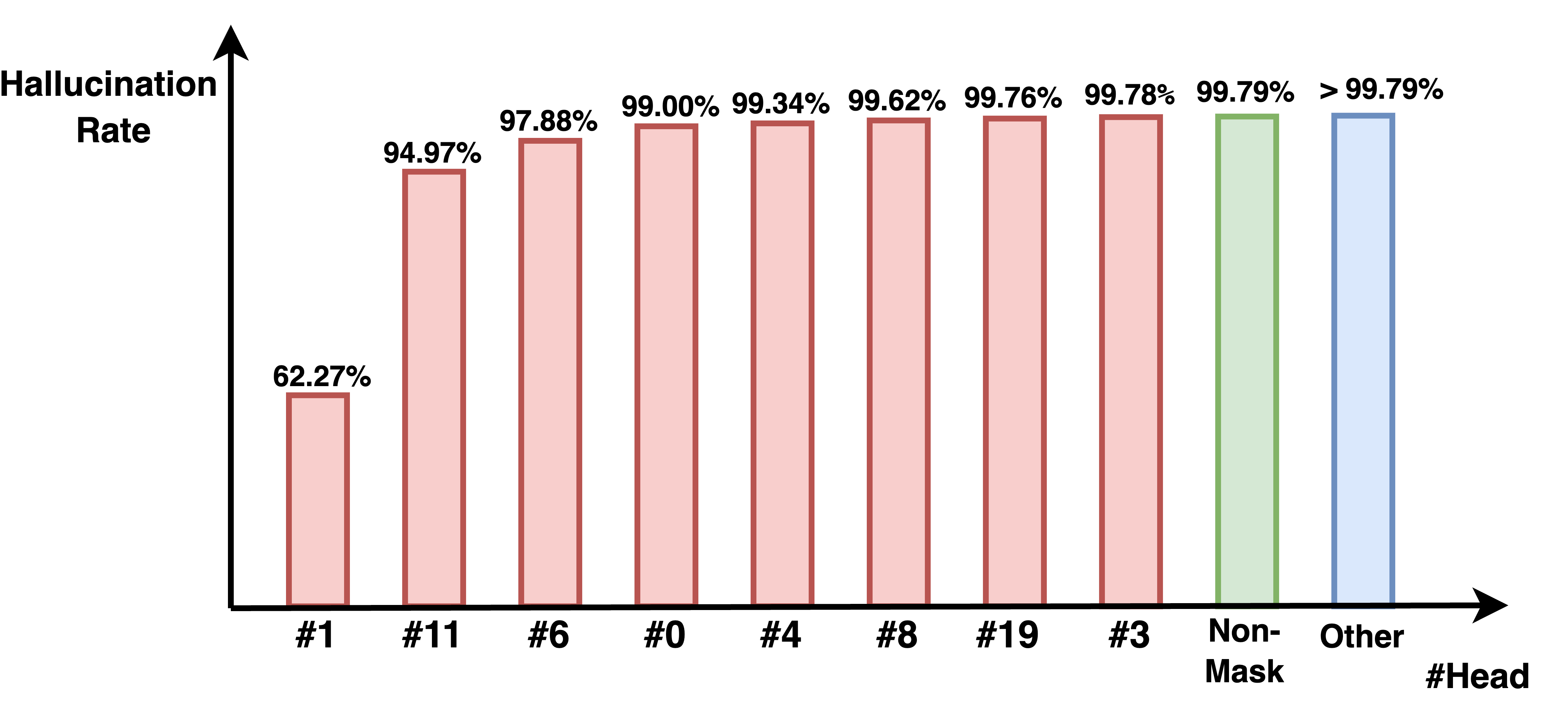}
  \caption{Whisper-large-v3's hallucination rate on UrbanSound8K with one particular self-attention head in the decoder masked.}
  \label{fig:content}
\end{figure}

The head-wise benchmark results are presented in Figure 2. Compared to the scenario without any masking (marked in green), masking any one of the following eight heads—\#0, \#1, \#3, \#4, \#6, \#8, \#11, or \#19—leads to varying degrees of reduction in the hallucination rate on the UrbanSound8K dataset (marked in red). This indicates that these heads contribute to hallucinations and are thus identified as hallucinatory heads. In contrast, masking one of the other 12 heads increases the hallucination rate (marked in blue), indicating their robustness in handling pure noises.

Notably, among the eight identified hallucinatory heads, masking head \#1 leads to a reduction of the hallucination rate by more than 30\%, demonstrating that head \#1 is the craziest head compared to the other heads.

\subsection{Multi-Head Masking}
After validating our hypothesis, we attempt to mask combinations of the hallucinatory heads to further minimize the hallucination rate. However, considering that the mask operation could negatively impact speech recognition performance, we also calculate the Word Error Rate (WER) on LibriSpeech \cite{librispeech} test-clean and test-other sets to illustrate the trade-off between reducing hallucinations and maintaining recognition accuracy.

\begin{table}[h!]
  \caption{Results of masking combinations of heads in Whisper-large-v3 decoder. The Hallucination Rate on UrbanSound8K and the Word Error Rate (WER) on LibriSpeech test-clean and test-other are reported.}
  \label{tab:hal_rate}
  \centering
  \resizebox{\columnwidth}{!}{
  \begin{tabular}{c c c c}
    \toprule
     Masked Heads & Hallucination Rate & WER test-clean & WER test-other \\
    \midrule
    Non-Mask & 99.97\% & \textbf{2.12\%} & \textbf{4.07\%} \\
    \#1, \#6 & 50.16\% & 5.70\% & 5.48\%\\
    \#1, \#11 & 70.03\% & 3.80\% & 6.07\%\\
    \#6, \#11 & 57.08\% & 2.37\% & 4.66\%\\
    \#1, \#6, \#11 & \textbf{24.10\%} & 3.57\% & 5.98\% \\
    \#0, \#1, \#6, \#11 & 28.91\% & 15.87\% & 21.39\%\\
    \bottomrule
  \end{tabular}
  }
\end{table}

First, following the order shown in Figure 2, we infer the model by masking each time two of the top three most hallucinatory heads: \#1, \#11, and \#6. 
Next, based on this, we further mask all three/four most hallucinatory heads.
As shown in Table 3, masking all the top three heads lowers the hallucination rate to a minimum of 24.10\%. Compared to the non-mask scenario, this highlights that the joint contribution of these three heads results in over 75\% of the hallucinations. Meanwhile, we observed that masking three heads, compared to masking four, makes little difference in transcription accuracy, with the WER on LibriSpeech test-clean increasing from 2.12\% to 3.57\% and test-other increasing from 4.07\% to 5.98\%.
These results suggest that masking three heads offers an optimal trade-off between hallucination mitigation and ASR accuracy, which serves as the experimenral groundwork for the calm-down fine-tuning in Section 4.
In addition, we found that the benefits of masking multiple heads were not as clear-cut as expected. For example, the hallucination rate increases when masking heads \#1 and \#11 together versus masking head \#1 alone, which points to a complex interplay among the heads.

\section{Calm-Down Fine-tuning}

Based on the findings in Section 3.2, we search to further reduce the hallucination rate while minimizing the loss of speech recognition accuracy. We consider that retaining all heads during inference, rather than masking them out directly, allows for a closer approximation to the original recognition performance. Therefore, we fine-tune the three most hallucinatory heads to guide the model to remain calm when processing non-speech inputs. As shown in Figure 3, we freeze all other parameters in the Whisper model, keeping only the parameters of heads \#1, \#6, and \#11 in each decoder layer trainable. Additionally, we train the model using purely noise data with blank labels.

\begin{figure}[t!]
  \centering
  \includegraphics[width=\linewidth]{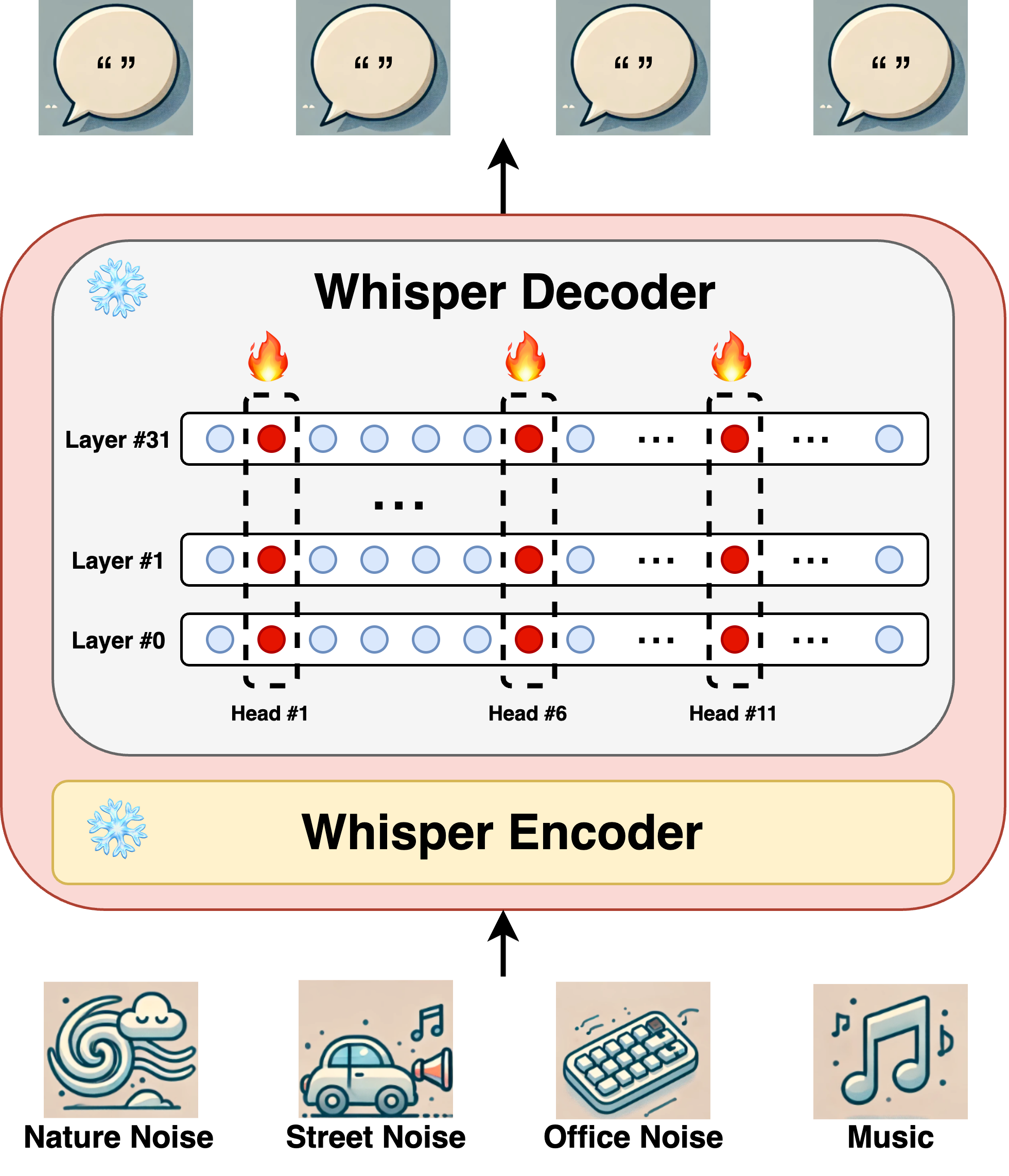}
  \caption{Calm-down fine-tuning designed to mitigate hallucinations in Whisper-large-v3. Specifically, all parameters in the Whisper-large-v3 model are frozen except for the parameters in head \#1, \#6, and \#11 of each decoder layer.  The model is trained using various non-speech noises paired with empty label strings to discourage it from reacting to non-speech inputs.}
  \label{fig:content}
\end{figure}

\subsection{Training set}
Our fine-tuning utilizes three datasets—AudioSet, DEMAND, and Musan—encompassing a total of 105 hours of non-speech data.

AudioSet \cite{AudioSet} is a comprehensive collection of audio events designed to support various tasks in audio analysis. It consists of over two million 10-second audio clips sourced from YouTube, labeled across 527 classes of sound events, including human speech, environmental sounds, music, and more. In our experiments, we utilize a subset of 11,753 audio clips, and exclude those segments labeled as "human speech".

The Diverse Environments Multichannel Acoustic Noise Database (DEMAND) \cite{demand} provides high-quality recordings of environmental noises. DEMAND features recordings from 18 different acoustic environments, captured using a 16-channel microphone array. These environments range from quiet indoor spaces to busy outdoor locations, offering a wide spectrum of real-world noise conditions. For audio files that exceed 30 seconds, we split them into segments of 30 seconds or shorter.

The MUSAN dataset \cite{musan} is a large-scale corpus designed for speech, music, and noise augmentation in speech processing tasks. The dataset covers a wide variety of audio types, including diverse music genres, natural and artificial noise sources, and speech recordings in multiple languages. Similarly, we only utilize the music and noise subsets, and chunk long audio files into 30-second clips.

For the calm-down fine-tuning we use only non-speech data, this has two key benefits: first, it minimizes the need for transcription, significantly reducing data annotation requirements. Second, by excluding speech data from the fine-tuning process, it ensures that the benchmark results on LibriSpeech remain unaffected in order to form a meaningful contrast, highlighting only the impact of non-speech fine-tuning on the WER of speech contents. 

\subsection{Experimental Settings}
We first conduct a contrast experiment by fine-tuning the entire decoder, aiming to assess the necessity of limiting fine-tuning to only three identified hallucinatory heads. We fine-tune the entire decoder on our training set for 3 epochs.

Next, for calm-down fine-tuning with the three most hallucinatory heads, we conduct experiments using different numbers of iterations. These experiments aim to observe the impact of varying fine-tuning depth on the model's overall performance. For each experiment, we fine-tune the model using a total batch size of 128 and a learning rate of $10^{-6}$ with a warm-up phase around 15\% of the total iterations. 

\subsection{Results and Discussion}
In the same manner as in Section 3.2, we benchmark the fine-tuned models on UrbanSound8K and the LibriSpeech test-clean and test-other datasets, comparing their performance with the non-fine-tuned models. It should be noted that we do not apply any head masks during the inference for the fine-tuned models. The results are shown in Table 3.

\begin{table}[h!]
  \caption{The hallucination rate on UrbanSound8K and the WER on LibriSpeech test-clean and test-other for non-fine-tuned models (first two rows) and calm-down fine-tuned models (last three rows).}
  \label{tab:hal_rate}
  \centering
  \resizebox{\columnwidth}{!}{
  \begin{tabular}{c c c c}
    \toprule
     Model & Hallucination Rate & WER test-clean & WER test-other \\
    \midrule
    Non-Mask & 99.97\% & \textbf{2.12\%} & \textbf{4.07\%} \\
    Mask\_\#1\_\#6\_\#11 & 24.10\% & 3.57\% & 5.98\% \\
    \midrule
    ft-decoder-3epochs & \textbf{0,01\%} & 100\% & 100\%\\
    ft-3heads-3epochs & 69.79\% & 2.16\% & 4.08\% \\
    ft-3heads-5epochs& 15.51\% & 2.19\% & 4.13\%\\
    \bottomrule
    \end{tabular}
}
\end{table}

For the non-fine-tuned models, we copy directly the results from Table 2, including the non-mask inference \textit{Non-Mask}, and the inference with 3 crazy heads masked \textit{Mask\_\#1\_\#6\_\#11}. For the fine-tuned models, we list three variations: (1) \textit{ft-decoder-3epochs}, where we fine-tune the entire decoder for three epochs; (2) \textit{ft-3heads-3epochs}, where we fine-tune only three crazy heads for three epochs; and (3) \textit{ft-3heads-5epochs}, where we fine-tune only three crazy heads for five epochs.

We first analyze whether it is necessary to perform fine-tuning only on crazy heads. The \textit{ft-decoder-3epochs} model reduces the hallucination rate to almost zero compared to the original model without fine-tuning. However, this improvement comes at a high cost, with the WER on Librispeech rising to 100\%, demonstrating that fine-tuning the entire decoder on non-speech data severely destroys the model's recognition performance.
In comparison, while reducing the hallucination rate from the initial value 99.97\% to 69.79\%, \textit{ft-3heads-3epochs} maintains remarkable stability on the LibriSpeech benchmarks under the same 3-epoch fine-tuning, with only a 0.04\% WER drop on the test-clean set and a 0.01\% drop on test-other. 
The contrast reveals that when only 3 crazy heads in the decoder are fine-tuned, the remaining non-fine-tuned heads in the decoder can preserve much of the model's original capabilities on speech recognition, offering necessary redundancy and robustness. In opposition, fine-tuning all the decoder heads can disrupt their established collaboration and overwrite the pre-trained knowledge, ultimately leading to a significant decline in performance.

Next, we find that the hallucination rate of \textit{ft-decoder-3epochs} remains high compared to the 24.10\% of \textit{Mask\_\#1\_\#6\_\#11}. Given the low WER on LibriSpeech benchmarks, there is still room to conduct further fine-tuning. As a result of deeper fine-tuning, the model \textit{ft-3heads-5epochs} achieves a better balance between hallucination and speech recognition accuracy, significantly reducing the hallucination rate to 15.51\%, with only minor increases in LibriSpeech WER: 0.07\% on test-clean and 0.06\% on test-other.
We also tracked the number of hallucinations that span more than five tokens and discovered a significant reduction from 742 to 91 after 5-epoch calm-down fine-tuning, illustrating the universal effectiveness of our method.

Additionally, we further investigate the impact of fine-tuning depth. Figure 4 illustrates the evolution of the progression of the hallucination rate and the WER on LibriSpeech test-clean as fine-tuning epochs increase. It can be observed that as fine-tuning becomes deeper, hallucination drops sharply at the beginning and then flattens, while the WER moves from slow increases to steeper rises. The best balance is reached between five and six epochs.
As a conservative measure, we determine that the model fine-tuned for 5 epochs is the most ideal, which we name as Calm-Whisper.
\begin{figure}[t!]
  \centering
  \includegraphics[width=\linewidth]{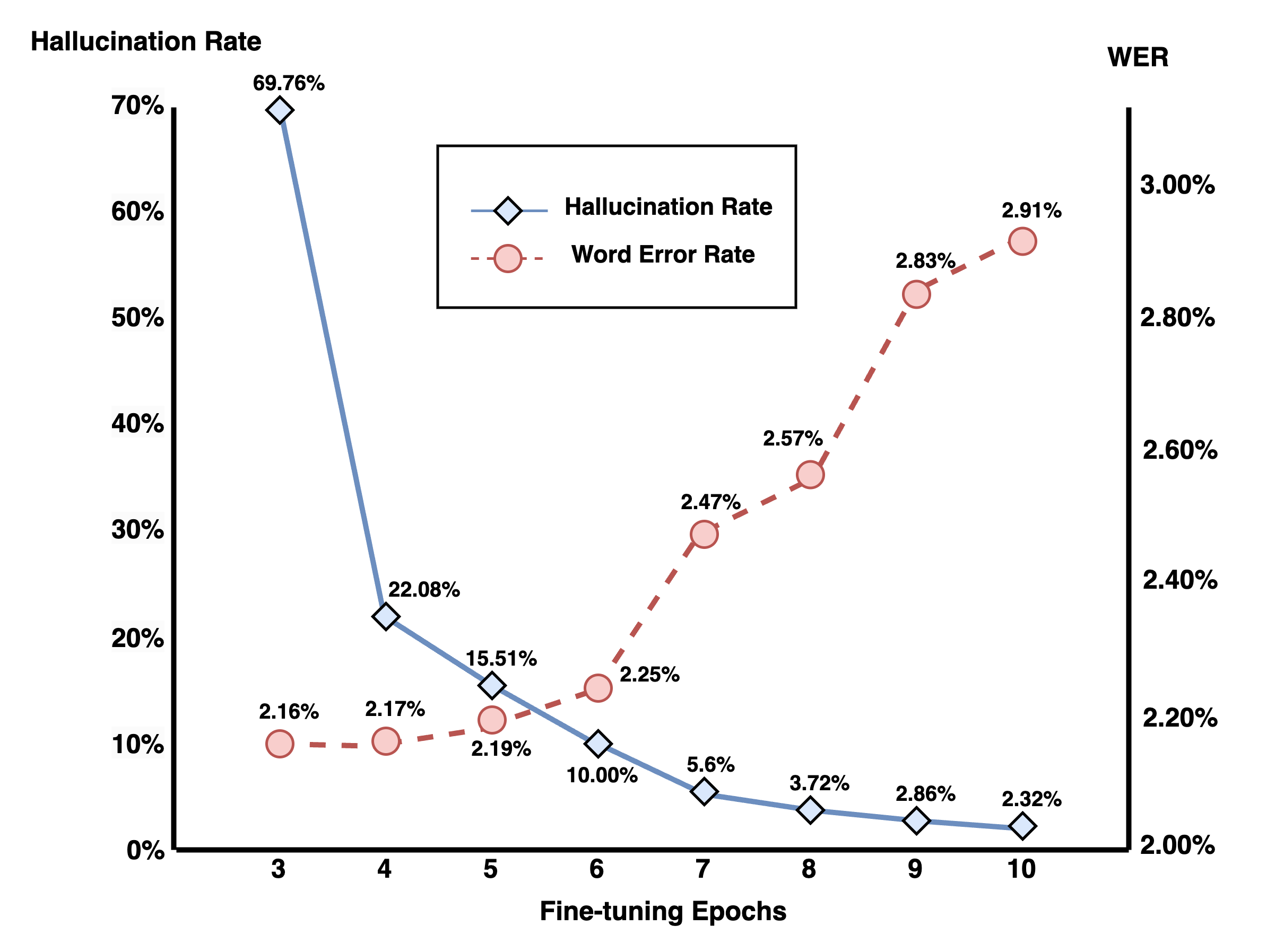}
  \caption{The evolution of the hallucination rate on UrbanSound8K and the WER on LibriSpeech test-clean across successive calm-down fine-tuning epochs for Whisper-large-v3.}
  \label{fig:content}
\end{figure}

\section{Conclusion}

In this paper, we investigate the hallucination issue of the Whisper model on non-speech segments. By applying a head-wise mask and benchmarking the hallucination rate, we identify the most problematic heads in the Whisper-large-v3 decoder that cause hallucination. We then perform calm-down fine-tuning which focuses on only the three craziest heads using collected non-speech data. Additionally, we explore the impact of different fine-tuning epochs to optimize the trade-off between hallucination and speech recognition accuracy. Through our work, we hope to help mitigate Whisper’s limitations in industrial applications and offer new perspectives on solving Whisper's hallucination issues.

\newpage

\bibliographystyle{IEEEtran}
\bibliography{mybib}

\end{document}